\documentclass[11pt]{article}

\usepackage[final]{acl}

\usepackage{times}
\usepackage{latexsym}

\usepackage[T1]{fontenc}

\usepackage[utf8]{inputenc}

\usepackage{microtype}

\usepackage{inconsolata}

\usepackage{graphicx}
\usepackage{amsmath}
\usepackage{amsfonts}
\usepackage{algorithm}
\usepackage{algpseudocode}
\usepackage{booktabs}
\usepackage{subcaption}
\usepackage{hyperref}

\usepackage{booktabs}
\usepackage{tabularx}
\usepackage{array}
\usepackage{ragged2e}

\usepackage{xcolor}
\usepackage{listings}
\usepackage{caption}

\newcommand{\hlcode}[1]{%
  \begingroup
  \setlength{\fboxsep}{1pt}%
  \fcolorbox{red}{white}{\strut\texttt{#1}}%
  \endgroup
}

\lstdefinelanguage{SPARQL}{
  morekeywords={PREFIX,SELECT,DISTINCT,WHERE,FILTER},
  sensitive=false,
  morecomment=[l]{\#},
  morestring=[b]"
}

\title{GGC: Selective Query Correction for Reliable Text-to-SPARQL Generation}

\author{
  \textbf{Yang Ziyi\textsuperscript{1}},
  \textbf{Thanh-Son Nguyen\textsuperscript{2}},
  \textbf{Lihui Chen\textsuperscript{1}}
\\
\\
  \textsuperscript{1}Nanyang Technological University,
  Centre for Info. Sciences and Systems
\\
  \textsuperscript{2}Institute of High Performance Computing,
\\
  Agency for Science, Technology and Research (A*STAR), Singapore
\\
\\
  \small{
    Yang Ziyi: \href{mailto:zyang025@e.ntu.edu.sg}{\texttt{zyang025@e.ntu.edu.sg}}
\\
    Thanh-Son Nguyen: \href{mailto:Nguyen_Thanh_Son@a-star.edu.sg}{\texttt{Nguyen\_Thanh\_Son@a-star.edu.sg}}
\\
    Lihui Chen: \href{mailto:ELHCHEN@ntu.edu.sg}{\texttt{ELHCHEN@ntu.edu.sg}}
  }
}

\author{
  \textbf{Yang Ziyi\textsuperscript{1}},
  \textbf{Thanh-Son Nguyen\textsuperscript{2}},
  \textbf{Nguyen Tuan Anh\textsuperscript{1}},
  \textbf{Lihui Chen\textsuperscript{1}}\thanks{
    Corresponding author: \href{mailto:elhchen@ntu.edu.sg}{\texttt{elhchen@ntu.edu.sg}}.
  }
\\
\\
  \textsuperscript{1}Nanyang Technological University,
  Centre for Info. Sciences and Systems
\\
  \textsuperscript{2}Institute of High Performance Computing,
\\
  Agency for Science, Technology and Research (A*STAR), Singapore
}

\begin{document}
\maketitle
\begin{abstract}
Large language models (LLMs) have demonstrated strong capabilities in structured query generation, making them a natural choice for Text-to-SPARQL, which translates natural language questions into executable SPARQL queries over knowledge graphs. However, their initial outputs remain unreliable: generated queries may be executable yet semantically misaligned with input questions, leading to incorrect retrieval. To address this issue, we propose Generator–Gate–Corrector (GGC), a framework for reliable LLM-based Text-to-SPARQL generation. GGC first uses a Generator to produce an initial query, then applies a Gate to predict whether correction is needed, and finally invokes a Corrector only for selected high-risk queries. This selective correction mechanism avoids unnecessary modifications and reduces the risk of degrading originally correct queries. Experiments on MCQA show that GGC improves query-level accuracy from 90.23\% to 98.33\% while reducing inference overhead by 45\% compared with correcting all generated queries. Ablation studies show that the Gate is robust across thresholds and that Corrector training data composition affects correction effectiveness and stability. Overall, the results demonstrate that selective correction enhances the accuracy, reliability, and efficiency of LLM-based text-to-SPARQL generation.

\end{abstract}

\section{Introduction}
Knowledge graph question answering (KGQA) aims to answer natural language questions (NLQs) by grounding them in the structured facts of knowledge graphs (KGs). Existing KGQA methods are typically categorized into semantic parsing (SP)-based methods and information retrieval (IR)-based methods \cite{9960856}. The former generates logical forms or query structures, while the latter obtains answers through entity, relation, or path retrieval over the graph.

In the context of RDF-based \cite{rdf12concepts} KGs, Text-to-SPARQL is an important approach to KGQA. RDF-based KGs represent knowledge as subject-predicate-object triples, and SPARQL \cite{sparql11query} is the standard query language for RDF data. Therefore, Text-to-SPARQL aims to translate NLQs into executable and semantically consistent SPARQL queries. Compared with direct natural language answer generation, this approach retrieves answers through explicit query execution over an external KG, making the intermediate query and retrieved graph facts inspectable and facilitating more verifiable and interpretable results \cite{pan2024unifying}.

In recent years, large language models (LLMs) have demonstrated strong capabilities in natural language understanding, generation, and code-related tasks \cite{10.5555/3495724.3495883,chen2021evaluating}. These capabilities make them a natural choice for Text-to-SPARQL, where models must understand user questions and generate semantically consistent and executable structured queries.

However, Text-to-SPARQL is not merely a text generation task. Generated queries must be executable while faithfully encoding the entities, relations, constraints, variable bindings, and reasoning paths expressed in the NLQ \cite{10.1145/3477495.3531841}. Since executability alone does not guarantee semantic correctness, LLM-based Text-to-SPARQL requires mechanisms to verify consistency with the original question, especially as LLMs may produce plausible but unfaithful outputs \cite{xu-etal-2023-fine}.

Our experiments further show that this problem is particularly prominent in Text-to-SPARQL. While the fine-tuned Generator achieves high query-level accuracy, many of its remaining errors are executable yet semantically flawed. We define Generator errors as initial SPARQL queries produced by the Generator that are judged incorrect against the ground-truth queries. On the Movie Complex Question Answering (MCQA) \cite{hoang2024semi} dataset, semantic errors account for 77.70\% of Generator errors, while syntactic errors account for only 21.07\%. This indicates that relying solely on execution failures to trigger corrections will miss many semantic errors.

The key insight of this paper is that in LLM-based Text-to-SPARQL, the issue is not only how to generate queries, but also when to correct them. Correcting all generated results may fix incorrect queries, but it also incurs higher inference overhead and may degrade originally correct queries. In contrast, selective correction concentrates computation on high-risk samples, improving accuracy while reducing unnecessary rewriting. Experimental results show that the proposed Generator--Gate--Corrector (GGC) framework improves query-level accuracy from 90.23\% to 98.33\%, with only about 10\% additional inference time over Generator-only inference. Compared with correcting all samples, selective correction achieves higher accuracy while reducing inference time by about 45\%.

In summary, our contributions are as follows:
\begin{itemize}
    \item We analyze LLM-based Text-to-SPARQL from a reliability perspective and show that executable but semantically flawed queries are a significant source of errors affecting system performance.
    \item We propose the GGC framework, which performs selective query correction through a detect-then-correct strategy and achieves a better balance between accuracy and inference efficiency.
    \item We conduct systematic experiments and ablation analyses on MCQA to examine the impact of the Generator, Gate, Corrector, KG execution feedback, and Corrector training data composition, providing empirical evidence for building more reliable Text-to-SPARQL systems. Additional results on SciQA \cite{Auer2023TheSS}, reported in Appendix \ref{app:cross-dataset}, further support these findings.
\end{itemize}

\section{Related Work}
\subsection{Knowledge Graph Question Answering}
Existing KGQA methods can generally be divided into SP-based methods and IR-based methods. SP-based methods convert natural language questions into logical forms, query graphs, or structured queries that can be executed on a knowledge graph to obtain answers \cite{9960856}. These methods provide relatively explicit reasoning processes, but their performance depends heavily on the quality of the generated intermediate structures. In contrast, IR-based methods retrieve relevant entities, relations, or subgraphs from topic entities and obtain answers through path search or neural reasoning. They are flexible for multi-hop reasoning but are often less interpretable than explicit query-based methods.

Query Graph Generation (QGG) \cite{lan-jiang-2020-query} and Neural State Machine-hybrid ($\mathrm{NSM}_{\mathrm{h}}$) \cite{10.1145/3437963.3441753} are representative methods of these two directions respectively. QGG constructs query graphs with entities, relations, constraints, and answer variables, while $\mathrm{NSM}_{\mathrm{h}}$ improves multi-hop reasoning by learning intermediate supervision signals. Unlike these methods, we focus on the reliability of LLM-based Text-to-SPARQL generation, especially how to identify and correct high-risk SPARQL queries after generation. Since QGG and $\mathrm{NSM}_{\mathrm{h}}$ are important reference baselines on the MCQA dataset, they are included in the experiments for comparison.

\subsection{Text-to-SPARQL and SPARQL Semantic Parsing}
With the development of pre-trained language models, Text-to-SPARQL has shifted from rule-based, template-based, and task-specific semantic parsing methods toward generation-based approaches \cite{9960856}. Prior work has compared BART, T5, and pointer-generator models on LC-QuAD 1.0 and LC-QuAD 2.0, showing that pre-trained models provide strong baselines for SPARQL semantic parsing \cite{10.1145/3477495.3531841}.

Recent studies have further applied LLMs to SPARQL generation \cite{xu-etal-2023-fine,dabramo-etal-2025-investigating} over Wikidata \cite{,10.1145/2629489}. By translating NLQs into executable SPARQL queries, these methods retrieve answers from KGs through an explicit query execution process, improving verifiability compared with direct answer generation. More recently, FIRESPARQL introduced a modular LLM-based framework for SPARQL generation over scholarly knowledge graphs \cite{kdir25}, combining fine-tuning, optional retrieval-augmented context, and query correction. These studies demonstrate the potential of LLMs for SPARQL generation, but they pay less attention to the reliability of generated queries, especially executable but semantically flawed queries. In contrast, our work focuses on when post-generation correction should be triggered. The proposed Generator–Gate–Corrector framework uses a learned Gate to selectively route only high-risk queries to the Corrector, reducing unnecessary rewriting and inference overhead.

\subsection{Constrained Generation and Error Correction in Structured Query Generation}
Text-to-SPARQL and Text-to-SQL are both natural-language-to-structured-query generation tasks that require syntactically valid and semantically correct outputs. Therefore, constrained decoding and error correction methods in Text-to-SQL provide useful references for Text-to-SPARQL.

For constrained decoding, PICARD \cite{scholak-etal-2021-picard} constrains autoregressive decoding through incremental parsing and rejects tokens that violate SQL syntax. For error correction, Text-to-SQL works \cite{chen-etal-2023-text,10.1145/3737873} have shown that post-generation correction can improve semantic parsing accuracy, and that structured clause-level editing is often more suitable than token-level editing for query repair.

These studies show that improving structured query generation does not only rely on stronger initial generators, but can also benefit from constraint, validation, and correction modules. 

\subsection{Motivation for Selective Correction in LLM-based Text-to-SPARQL }
Existing work has advanced KGQA through graph reasoning, semantic parsing, and structured query generation. However, LLM-based Text-to-SPARQL methods still face reliability issues, where executable queries may be semantically inconsistent with the original question. Inspired by recent post-correction studies in Text-to-SQL, we explore post-correction for LLM-based Text-to-SPARQL. Unlike prior work focusing mainly on query generation, our approach emphasizes post-generation error detection and selective correction to determine when and how corrections should be applied for improved semantic consistency.

\section{Methodology}
We propose a Generator–Gate–Corrector framework to improve the reliability of LLM-based Text-to-SPARQL generation. Instead of correcting all generated queries, this framework first determines whether the initial SPARQL query is likely to contain errors, and then triggers corrections only for high-risk queries. The overall goal is to improve query accuracy while reducing unnecessary correction and additional inference overhead.

\subsection{Task Formulation}
Given an NLQ, $q$, the objective of Text-to-SPARQL is to generate an executable SPARQL query, $s$. The answer set can be retrieved from the KG by executing the query. Compared with general text generation tasks, Text-to-SPARQL requires the generated queries to satisfy the following requirements: First, queries must conform to the syntax of SPARQL, and second, queries must be semantically consistent with original questions.

The initial query is generated by the Generator, denoted as $s^{(0)}$. If the query is determined to have potential errors, a corrected query, $s^{(c)}$, is generated by the Corrector. The final output query is denoted as $\hat{s}$. Therefore, we focus on how to generate $s^{(0)}$ and on how to determine whether it needs to be corrected.

\begin{figure*}[t]
    \centering
    \includegraphics[width=\textwidth]{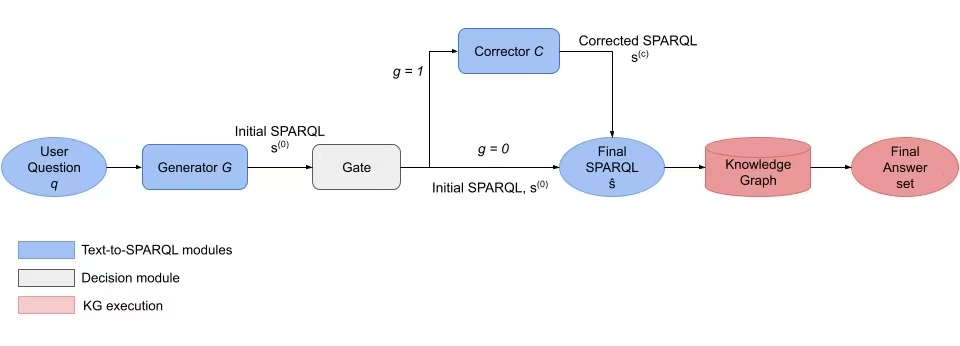}
    \caption{The proposed Generator-Gate-Corrector framework for Text-to-SPARQL generation.}
    \label{fig:ggc-framework}
\end{figure*}

\subsection{Generator-Gate-Corrector Framework}
We separate the Text-to-SPARQL process into three stages: initial generation, error detection, and selective correction. The overall process is as follows:
\[s^{(0)} = G(q)\]
\[g = Gate(q, s^{(0)})\]
\[
\hat{s} =
\begin{cases}
s^{(0)}, & \text{if } g = 0 \\
C(q, s^{(0)}), & \text{if } g = 1
\end{cases}
\]
where $G(\cdot)$ denotes the Generator, $C(\cdot)$ denotes the Corrector, and $g \in \{0,1\}$ is the binary output of the Gate. The framework is shown in Figure \ref{fig:ggc-framework}.

Since not all initial SPARQL queries need to be corrected, triggering the Corrector on all queries would significantly increase inference overhead and could also corrupt queries that were initially correct. Therefore, the Gate in the framework acts as a query risk assessor, ensuring that the Corrector is triggered only on high-risk queries.

\subsection{Generator}
The objective of the Generator is to translate an NLQ, $q$, into an initial SPARQL query, $s^{(0)}$. We use a supervised fine-tuned LLM as the Generator. During training, the input is the NLQ, and the target output is the corresponding ground-truth SPARQL query. During inference, the Generator generates an initial query $s^{(0)}$ based on the input question.

The Generator aims to produce an initial candidate query that is as accurate as possible. However, due to SPARQL's strict requirements for entities, relations, constraints, and variable bindings, the Generator's output may still contain syntactic or semantic errors. Therefore, the initial query is not directly considered the final result but is further passed to the Gate for reliability assessment.

\subsection{Gate}
The Gate is a binary classification module used to determine whether the initial SPARQL query needs correction. Its input consists of an NLQ and the initial query produced by the Generator:
$$x_{gate} = [q;s^{(0)}]$$

The Gate outputs the probability that correction is required, and the final decision is obtained based on the threshold $\tau$:
$$
g=\mathbb{I}\left[P\left(g=1 \mid x_{gate}\right) \geq \tau\right]
$$
where, $g=1$ indicates that the Corrector is triggered, and $g=0$ indicates that the initial SPARQL query is kept. The default threshold is $\tau = 0.5$.

Compared with methods that rely solely on execution feedback, the objective of the Gate is to identify queries that may contain syntactic or semantic errors, including executable queries that are semantically inconsistent with the NLQ. This is important because many incorrect SPARQL queries are executable but return answers that do not match the question. Therefore, the Gate needs to identify not only syntactic or execution-related errors, but also potential semantic errors.

\subsection{Corrector}
The Corrector aims to revise SPARQL queries identified as high-risk by the Gate. Unlike the Generator, the input to the Corrector contains an NLQ and a corresponding initial SPARQL query.

The objective of the Corrector is to fix potential errors in entities, relations, constraints, variable bindings, or query structure while preserving the correct parts of the initial query. During training, the Corrector takes an NLQ and an initial query produced by the Generator as input, and the corresponding ground-truth SPARQL query as the target output.

We further consider different training data compositions for the Corrector. Using only incorrect samples can provide a more direct correction signal, while adding a certain proportion of originally correct samples can help the Corrector learn to maintain stable output when the query is already correct. The impact of different training data compositions on correction effectiveness and stability will be examined in section 4.3.

\subsection{Training and Inference}
Both the Generator and the Corrector are trained using standard supervised fine-tuning, and training samples are constructed using an instruction-response format. For the Generator, the instruction includes a task description and an NLQ, while the response is a ground-truth SPARQL query. For the Corrector, the instruction includes a task description, an NLQ, and an initial SPARQL query, while the response is a ground-truth SPARQL query. During training, the language modeling loss is applied only to the response portion.

The Gate is trained as a separate binary classifier. Its training samples consist of an NLQ, an initial SPARQL query, and a binary label. If the initial query is incorrect, the label is "Correction Required". If the initial query is correct, the label is "No Correction Required". 

The inference procedure is summarized in Algorithm \ref{alg:ggc-inference}:
\begin{algorithm}[!htbp]
\caption{Generator-Gate-Corrector Inference}
\label{alg:ggc-inference}
\begin{algorithmic}[1]
\Require Natural language question $q$
\Ensure Final SPARQL query $\hat{s}$

\State $s^{(0)} \leftarrow G(q)$
\State $g \leftarrow \mathrm{Gate}(q, s^{(0)})$
\If{$g = 1$}
    \State $s^{(c)} \leftarrow C(q, s^{(0)})$
    \State $\hat{s} \leftarrow s^{(c)}$
\Else
    \State $\hat{s} \leftarrow s^{(0)}$
\EndIf
\State \Return $\hat{s}$

\end{algorithmic}
\end{algorithm}

Through this pipeline, the system selectively corrects high-risk queries, achieving a better balance between accuracy, reliability, and inference efficiency.
\section{Experiments and Results}
This section first introduces the experimental setup and compared systems, then presents the end-to-end framework results and ablation studies on key components, including the Generator, Gate and its threshold sensitivity, KG execution feedback, and Corrector training data composition.
\subsection{Experimental Setup}
\textbf{Dataset}
The experiments were conducted on the MCQA dataset \cite{hoang2024semi}. MCQA is a complex KGQA dataset for the movie domain constructed over iMKG, a KG based on Wikidata and MovieKG. Each sample contains an NLQ, question type, topic entity, answer, and corresponding ground-truth SPARQL query. This dataset is built on a movie-domain KG and is suitable for evaluating Text-to-SPARQL and KGQA methods.

The dataset includes training, validation, and test sets. The training set contains 119,409 samples, the validation set contains 15,785 samples, and the test set contains 31,570 samples. We use a portion of the training data to fine-tune the Generator, enabling it to learn the mapping from NLQs to SPARQL queries. Then, the trained Generator is used to generate initial queries on the remaining training data, and training samples for the Gate and Corrector are constructed accordingly. If the initial query generated by the Generator is incorrect, the corresponding sample is labeled "Correction Required". Otherwise, it is labeled "No Correction Required". 

\begin{table}[t]
\centering
\small
\setlength{\tabcolsep}{4pt}
\renewcommand{\arraystretch}{1.12}
\begin{tabular}{@{}p{0.38\columnwidth}p{0.60\columnwidth}@{}}
\toprule
\textbf{Setting} & \textbf{Description} \\
\midrule
Generator-only
& No correction; output $s^{(0)}$. \\

Gen. + Corr. 
& Correct all initial queries. \\

Gen. + Gate + Corr. 
& Selectively correct initial queries. \\

$\mathrm{NSM}_{\mathrm{h}}$ 
& IR baseline reported in MCQA. \\

QGG 
& SP baseline reported in MCQA. \\
\bottomrule
\end{tabular}
\caption{Evaluated settings and baselines. Gen. denotes the Generator, Corr. denotes the Corrector.}
\label{tab:settings_compared}
\end{table}

\begin{table*}[t]
\centering
\small
\setlength{\tabcolsep}{3pt}
\renewcommand{\arraystretch}{1.08}
\begin{tabular}{lccccc}
\toprule
\textbf{System} 
& \textbf{Query Acc. (\%)} 
& \textbf{Item Prec. (\%)} 
& \textbf{Item Rec. (\%)} 
& \textbf{Item F1 (\%)} 
& \textbf{Infer. Time (Total / Q.)} \\
\midrule
$\mathrm{NSM}_{\mathrm{h}}$ 
& -- 
& 59.20 
& 70.06 
& 59.44 
& -- \\

QGG 
& -- 
& 11.65 
& 25.13 
& 12.32 
& -- \\

Generator-only 
& 90.23 
& 37.96 
& 85.76 
& 52.63 
& $\sim$30h / $\sim$3.4s \\

Gen. + Corr. (correct-all)
& 92.34 
& 91.62 
& 93.43 
& 92.51 
& $\sim$60h / $\sim$6.8s \\

Gen. + Gate + Corr. (selective)
& \textbf{ 98.33 }
& \textbf{ 99.62 }
& \textbf{ 98.70 }
& \textbf{ 99.16 }
& $\sim$33h / $\sim$3.7s \\
\bottomrule
\end{tabular}
\caption{Main results on MCQA. The proposed selective correction framework achieves the best query-level accuracy while requiring much less inference time than the correct-all baseline. Gen. denotes the Generator and Corr. denotes the Corrector. Infer.Time (Total/Q.) refers to both the total inference time on the full test set, and the average time per test question, respectively.}
\label{tab:end-to-end-results}
\end{table*}
\noindent\textbf{Compared Settings} Table \ref{tab:settings_compared} summarizes the compared settings in the experiments. Generator-only evaluates the initial generation capability, while Gen. + Corr. tests the effect of correcting all initial queries. Gen. + Gate + Corr. represents the proposed selective correction framework. We also include $\mathrm{NSM}_{\mathrm{h}}$ and QGG as traditional KGQA reference baselines to assess end-task answer retrieval performance. Since these methods do not directly generate SPARQL queries or report query-level accuracy, their results are not strictly comparable to ours at the query level. Existing LLM-based Text-to-SPARQL methods are not included as main baselines because standardized implementations and evaluation settings remain limited.

\noindent \textbf{Implementation Details}
Both the Generator and Corrector are supervised fine-tuned based on Llama-3.2-3B-Instruct \cite{grattafiori2024llama}, employing LoRA \cite{hu2022lora} and 4-bit quantized loading \cite{10.5555/3666122.3666563} to reduce training costs. Both used an instruction-response format to construct training samples and applied the language modeling loss only in the response portion. 

The Gate is modeled as a binary classifier. We compare the RoBERTa-based \cite{liu2019roberta} Gate and the SBERT-based \cite{reimers-gurevych-2019-sentence, 10.5555/3495724.3496209} Gate, and use the RoBERTa-based  Gate, which performed better, as the default setting.

\noindent \textbf{Evaluation Metrics}
For the Generator, Corrector, and framework, we report both query-level and item-level metrics. Query-level accuracy measures the matching between the answer sets retrieved with generated SPARQL and with the ground-truth SPARQL, and is the primary evaluation metric. Item-level precision, recall, and F1 score measure the matching between the answer set obtained by the generated SPARQL and the ground-truth answer set, and are used for comparison with traditional KGQA baselines.

For the Gate, we report accuracy, precision, recall, F1 score, false positive rate (FPR), and trigger rate. "Correction Required" is considered the positive class, and the trigger rate represents the proportion of samples that the Gate determines need to proceed to the Corrector.

The detailed experimental setup and prompt templates are summarized in Appendices~\ref{app:detailed exp setup} and \ref{app:prompt design}.

\subsection{End-to-End Framework Results}
We evaluate the end-to-end GGC framework and compare it with traditional KGQA baselines on MCQA. The results are shown in Table~\ref{tab:end-to-end-results}. 

The proposed framework achieves strong overall performance, reaching 98.33\% query-level accuracy and 99.16\% item-level F1. Compared with traditional KGQA baselines, GGC obtains better item-level performance, showing that explicit SPARQL generation combined with selective correction can improve final answer retrieval quality.

However, this comparison should be interpreted carefully because the baselines and our method optimize different objectives. Traditional KGQA baselines such as $\mathrm{NSM}_{\mathrm{h}}$ and QGG are designed for answer retrieval, while our framework explicitly generates executable SPARQL queries. Since these baselines do not report query-level accuracy, they cannot be directly compared in terms of SPARQL generation quality. Therefore, we use them as reference baselines for MCQA answer retrieval, rather than as direct Text-to-SPARQL generation baselines.

Overall, the results show that the proposed GGC framework achieves competitive answer retrieval performance by additionally producing explicit SPARQL queries, improving accuracy, reliability, and inference efficiency.

\subsection{Ablation Study on the Impact of each Component}
To understand how each component contributes to the framework, we conduct ablation studies on them. We analyze the Generator-only performance, evaluate the Gate’s ability to identify high-risk queries, and study key design choices including Gate threshold, KG execution feedback, and Corrector training data composition.

\noindent\textbf{Generator-only Performance}
Generator-only achieves a query-level accuracy of 90.23\%, which indicates that the fine-tuned LLM already has a considerable Text-to-SPARQL generation capability. However, the item-level precision is only 37.96\%, significantly lower than the recall of 85.76\%. These results in Table~\ref{tab:end-to-end-results} indicate that although some generated queries may retrieve the correct answers, they also return a large number of irrelevant answers. In other words, Generator errors are not primarily non-executable queries, but are more likely to be semantic errors such as overly broad query scope, incorrect relation selection, missing constraints, or inaccurate variable binding.

To further analyze the error source of the Generator, we categorized the incorrect samples into syntax error, semantic error, and ground-truth SPARQL error. The results are shown in Table~\ref{tab:generator-error-distribution}. As can be seen, semantic errors account for 77.70\% of the Generator's errors, significantly higher than the 21.07\% for syntax errors. This indicates that the main challenge of LLM-based Text-to-SPARQL is not merely generating executable queries, but rather generating queries that are semantically consistent with the NLQs. Therefore, relying solely on execution failure to trigger correction is insufficient, as this approach misses a large number of executable but semantically incorrect queries.

This result supports the necessity of introducing the Gate. The goal of the Gate is to determine whether a query can be executed, as well as to determine whether the initial query may have semantic inconsistencies, thereby deciding whether the Corrector should be triggered.
\begin{table}[h]
\centering
\small
\setlength{\tabcolsep}{8pt}
\renewcommand{\arraystretch}{1.08}
\begin{tabular}{lcc}
\toprule
\textbf{Error Type} & \textbf{Count} & \textbf{Percentage (\%)} \\
\midrule
Syntax Error & 650 & 21.07 \\
Semantic Error & 2,397 & 77.70 \\
Gold SPARQL Error & 38 & 1.23 \\
\midrule
Total & 3,085 & -- \\
\bottomrule
\end{tabular}
\caption{Error distribution of the Generator.}
\label{tab:generator-error-distribution}
\end{table}

\noindent\textbf{Gate Performance}
To evaluate whether the Gate can identify which queries require correction, we compare the Gate with two backbones, and the results are shown in Table~\ref{tab:gate-model-comparison}.

Both backbones achieved high classification performance, indicating that strong error signals were already present between the NLQ and the initial SPARQL query. The RoBERTa-based gate is better than the SBERT-based gate in recall and F1 score. RoBERTa's recall reached 92.11\%, meaning it could cover most queries that truly needed correction. Meanwhile, its FPR was only 0.31\%, indicating it rarely misclassified originally correct queries.

This is particularly important for the entire pipeline. If the Gate's recall is too low, the Corrector will not be triggered for many incorrect queries. If the FPR is too high, the Corrector may be frequently triggered, corrupting originally correct queries. Therefore, the Gate's role is not only to reduce inference overhead but also to perform risk screening between correct and incorrect queries. Due to the superior overall performance of the RoBERTa-based gate, we use RoBERTa as the default Gate backbone in subsequent experiments.
\begin{table*}[t]
\centering
\small
\setlength{\tabcolsep}{10pt}
\renewcommand{\arraystretch}{1.08}
\begin{tabular}{lcccccc}
\toprule
\textbf{Gate Model} 
& \textbf{Acc. (\%)} 
& \textbf{Prec. (\%)} 
& \textbf{Rec. (\%)} 
& \textbf{F1 (\%)} 
& \textbf{FPR (\%)} 
& \textbf{Trigger Rate (\%)} \\
\midrule
RoBERTa 
& 98.95 
& 97.00 
& 92.11 
& 94.49 
& 0.31 
& 9.31 \\

SBERT 
& 98.32 
& 96.45 
& 86.03 
& 90.94 
& 0.34 
& 8.74 \\
\bottomrule
\end{tabular}
\caption{Comparison of Gate models.}
\label{tab:gate-model-comparison}
\end{table*}

\noindent\textbf{Effect of Selective Correction}
We further compare selective correction with Generator-only and Generator + Corrector, where the Corrector is applied to all generated queries. As shown in Table~\ref{tab:end-to-end-results}, Generator-only achieves 90.23\% query-level accuracy with approximately 30 hours of inference time. Correcting all samples increases the inference time to approximately 60 hours, but only improves query-level accuracy to 92.34\%.

In contrast, the complete GGC framework achieves 98.33\% query-level accuracy with approximately 33 hours of inference time. This indicates that the Gate is essential not only for reducing computational cost, but also for avoiding unnecessary correction of originally correct queries.

\noindent \textbf{Gate Threshold Sensitivity}
We analyze whether the Gate depends on careful threshold tuning. As shown in Table~\ref{tab:gate-threshold}, varying the threshold mainly introduces a small precision--recall trade-off: lower thresholds trigger more corrections and slightly improve recall, while higher thresholds reduce false triggers and slightly improve precision. However, the overall performance remains stable. 
\begin{table*}[!t]
\centering
\small

\begin{subtable}{\textwidth}
\centering
\begin{tabular}{ccccccc}
\toprule
\textbf{Threshold} 
& \textbf{Acc. (\%)} 
& \textbf{Prec. (\%)} 
& \textbf{Rec. (\%)} 
& \textbf{F1 (\%)} 
& \textbf{FPR (\%)} 
& \textbf{Trigger Rate (\%)} \\
\midrule
0.25 & 98.89 & 95.98 & 92.57 & 94.24 & 0.42 & 9.45 \\
0.50 & 98.95 & 97.00 & 92.11 & 94.49 & 0.31 & 9.31 \\
0.75 & 98.93 & 97.13 & 91.79 & 94.38 & 0.29 & 9.26 \\
\bottomrule
\end{tabular}
\caption{Effect of Gate threshold.}
\label{tab:gate-threshold}
\end{subtable}

\vspace{0.8em}

\begin{subtable}{\textwidth}
\centering
\begin{tabular}{lcccc}
\toprule
\textbf{Setting} 
& \textbf{Gate Acc. (\%)} 
& \textbf{Gate F1 (\%)} 
& \textbf{Pipeline Query Acc. (\%)} 
& \textbf{Pipeline Item F1 (\%)} \\
\midrule
With feedback & 99.36 & 96.66 & 98.70 & 99.09 \\
Without feedback & 98.95 & 94.49 & 98.27 & 97.72 \\
\bottomrule
\end{tabular}
\caption{Effect of KG feedback.}
\label{tab:kg-feedback}
\end{subtable}

\vspace{0.8em}

\begin{subtable}{\textwidth}
\centering
\begin{tabular}{lcccc}
\toprule
\textbf{Composition} 
& \textbf{Query Acc. (\%)} 
& \textbf{Item Prec. (\%)} 
& \textbf{Item Rec. (\%)} 
& \textbf{Item F1 (\%)} \\
\midrule
W1.0 & 98.27 & 97.09 & 98.37 & 97.72 \\
W1.0 + C0.2 & 98.33 & 99.62 & 98.70 & 99.16 \\
W1.0 + C0.5 & 98.63 & 87.66 & 98.85 & 92.92 \\
W1.0 + C0.8 & 98.80 & 99.66 & 99.00 & 99.33 \\
\bottomrule
\end{tabular}
\caption{Effect of Corrector training data composition.}
\label{tab:corrector-composition}
\end{subtable}

\caption{Ablation results of the framework.}
\label{tab:ablation-results}
\end{table*}

When the threshold changes from 0.25 to 0.75, the trigger rate only changes from 9.45\% to 9.26\%, and the F1 score remains around 94\%. This suggests that the Gate learns stable question-query mismatch signals rather than relying on a carefully tuned threshold. We therefore use $\tau = 0.5$ as the default setting.

\noindent\textbf{Effect of KG Execution Feedback}
 Intuitively, feedback from KG execution can provide useful information, such as whether the query can be executed successfully or whether the returned result is abnormal. Therefore, we compare the with-feedback and without-feedback settings for both the Gate and the pipeline.

The results in Table~\ref{tab:kg-feedback} demonstrate that KG feedback improves both Gate performance and end-to-end pipeline performance. For example, pipeline query-level accuracy increases from 98.27\% to 98.70\%. However, the improvement is relatively limited, especially on the main query-level metric. In contrast, the without-feedback setting avoids an additional execution step before correction, making the pipeline simpler and more efficient.

Therefore, KG feedback is useful but not essential. The strong without-feedback results suggest that the NLQ-SPARQL pairs already contain sufficient signals for detecting semantic errors. Considering both performance and efficiency, the without-feedback setting is more practical as the default configuration.

\noindent\textbf{Effect of Corrector Training Data Composition}
Finally, we analyze how adding correct samples affects Corrector training. We use $W$ to denote initially wrong queries produced by the Generator and $C$ to denote initially correct queries. $W1.0$ denotes training only on the full set of wrong samples, while $W1.0+C0.2$, $W1.0+C0.5$, and $W1.0+C0.8$ additionally add correct samples whose sizes correspond to 20\%, 50\%, and 80\% of all training samples, respectively. Thus, the training set size increases as more correct samples are added.

As shown in Table~\ref{tab:corrector-composition}, adding correct samples does not weaken the Corrector. Query-level accuracy increases from 98.27\% under $W1.0$ to 98.80\% under $W1.0+C0.8$, suggesting that correct samples help the model preserve original correct queries and reduce unnecessary rewriting. Since larger $C$ ratios also increase training overhead, we use $W1.0+C0.2$ as the default setting in the main experiments to balance performance and training efficiency.

Item-level metrics show some fluctuation. For example, $W1.0+C0.5$ achieves higher query-level accuracy than $W1.0+C0.2$, but its item-level precision and F1 are lower, possibly due to outlier queries with large answer sets. Therefore, we treat query-level accuracy as the primary metric and item-level metrics as supplementary evidence.

Overall, this ablation suggests that Corrector training benefits from both repair and preservation signals: wrong samples teach correction, while correct samples discourage unnecessary rewriting.

Additional case studies, subset-level Gate-Corrector coordination analysis and additional dataset preliminary results are provided in Appendices \ref{app:case study}, \ref{app:coordination}, and \ref{app:cross-dataset}.
\section{Conclusion}
In this paper, we presented a GGC framework for reliable LLM-based Text-to-SPARQL generation. The framework first generates an initial SPARQL query, uses a Gate to decide whether correction is needed, and triggers the Corrector only on high-risk queries, improving accuracy and reliability while reducing inference overhead.

Experiments on the MCQA dataset show that the main errors of the Generator are executable but semantically inconsistent queries. This suggests that execution failure alone is insufficient for triggering correction. The Gate identifies queries requiring correction with a low FPR, and the full framework substantially improves query-level accuracy over both Generator-only and correct-all settings. Compared with correcting all generated queries, selective correction achieves higher accuracy while reducing inference time by about 45\%.

Overall, reliable Text-to-SPARQL requires both strong generation capability and effective detection of when to correct. The proposed GGC framework demonstrates the potential of addressing this issue through a post-hoc and selective correction strategy, and we hope it will encourage further research in this direction.

\newpage
\section*{Limitations}
While this paper proposes a Generator–Gate–Corrector framework for reliable LLM-based Text-to-SPARQL generation and demonstrates its effectiveness on MCQA, we acknowledge several limitations that warrant further exploration in future work:
\begin{itemize}
    \item The main experiments are conducted on MCQA, a movie-domain dataset with label-based SPARQL queries. Although the results demonstrate the effectiveness of selective correction in this setting, evaluation beyond MCQA remains limited in scope. We provide an additional preliminary evaluation on another Text-to-SPARQL dataset in Appendix~\ref{app:cross-dataset}, but more comprehensive experiments across diverse domains, knowledge graphs, SPARQL formats, and model backbones are needed to further assess the generalizability of the GGC framework. In particular, extending the framework to ID-based SPARQL settings may require additional entity linking, schema alignment, and relation grounding mechanisms.
    \item Dependence on the Generator’s error distribution. The Gate and Corrector are trained using the outputs of a specific Generator. Therefore, their performance may depend on the error patterns of that Generator. When a different or stronger base model is used, the Gate and Corrector may need to be retrained or adapted.
    \item Additional offline computation. Although the proposed framework reduces unnecessary correction during inference, constructing training data for the Gate and Corrector still requires generating initial SPARQL queries on a large number of training samples. This introduces additional offline computational cost, especially when scaling to larger datasets or stronger base models.
\end{itemize}

\newpage
\section*{Ethical Considerations}
Our proposed framework focuses on improving the reliability of LLM-based Text-to-SPARQL generation over knowledge graphs. The experiments are conducted on an existing KGQA dataset and do not involve collecting private user data or annotating sensitive personal information. However, if such systems are deployed in real-world applications, incorrectly generated SPARQL queries may still lead to misleading or incomplete answers, especially when the underlying knowledge graph contains outdated, biased, or incomplete facts. Therefore, practical deployment should include appropriate validation, uncertainty indication, and human oversight in high-stakes scenarios. In addition, extending the framework to open-domain or ID-based knowledge graphs may introduce risks from entity linking errors and biases encoded in the knowledge graph, which should be carefully examined in future work. 

\section*{Acknowledgment}
This project was supported by Nanyang Technological University under the URECA Undergraduate Research Programme.

\newpage
\bibliography{custom}
\clearpage

\appendix

\section{Detailed Experimental Setup}
\label{app:detailed exp setup}
\subsection{Models}
The Generator and the Corrector are fine-tuned based on “unsloth/Llama-3.2-3B-Instruct" \cite{unsloth_llama32_3b_instruct}. The model has the following parameter counts: 
\begin{itemize}
    \item Total Parameters: 3,237,063,680
    \item Trainable Parameters: 24,313,586
\end{itemize}

The backbones of the Gate are implemented based on "FacebookAI/roberta-base" \cite{huggingface2024robertabase} and "sentence-transformers/all-MiniLM-L6-v2" \cite{sentenceTransformersAllMiniLML6v2}. These models have the following parameter counts:
\begin{itemize}
    \item RoBERTa-base: $\sim$ 125,000,000
    \item all-MiniLM-L6-v2: 22,713,216
\end{itemize}

\subsection{Hyper-parameters}
\textbf{Generator and Corrector}
The hyper-parameters of the Generator and the Corrector are as follows:
\begin{itemize}
    \item Maximum sequence length: 2048
	\item Batch size: 2
	\item Gradient accumulation steps: 4
	\item Number of epochs: 1
	\item Learning rate: $2 \times 10^{-4}$
	\item LoRA rank r: 16
	\item LoRA alpha: 16
	\item LoRA dropout: 0
	\item Optimizer: AdamW\_8bit \cite{loshchilov2018decoupled}
	\item Weight Decay: 0.01
    \item Random seed: 3407
\end{itemize}

\noindent\textbf{Gate}
The hyper-parameters of the Gate are as follows:
\begin{itemize}
    \item Maximum sequence length: 256
	\item Batch size: 8
	\item Number of epochs: 3
	\item Learning rate: $2 \times 10^{-5}$
    \item Random seed: 42
\end{itemize}

\subsection{Hardware and Software Environment}
All experiments were conducted on a server with the following configurations:
\begin{itemize}
    \item GPU: NVIDIA GeForce RTX 3090 with 24 GB VRAM.
    \item CPU: Intel Xeon W-2295
    \item Memory: 256 GB
    \item Operating System: Linux Ubuntu 22.04
    \item Deep learning framework: PyTorch 2.8.0 with CUDA 12.8.
    \item LLM fine-tuning libraries: Transformers 4.55.4, Datasets 3.6.0, Unsloth 2025.9.7
    \item Machine learning library: scikit-learn 1.8.0.
    \item Knowledge graph and SPARQL tools: RDFLib 7.2.1 and LangChain 0.3.27.
    \item Knowledge graph engine: GraphDB with a local SPARQL endpoint.
\end{itemize}

\subsection{Training Time}
Table \ref{tab:training-cost} summarizes the training time of the Generator, the Gates with two backbones, and the Correctors with different training data compositions.
\begin{table}[h]
\centering
\small
\setlength{\tabcolsep}{4pt}
\begin{tabular}{lccc}
\hline
\textbf{Setting} & \textbf{Training Samples} & \textbf{Epochs} & \textbf{Time} \\
\hline
Generator & 12,450 & 1 & $\sim$1h \\
Gate-RoBERTa & 88,506 & 3 & $\sim$1h \\
Gate-SBERT & 88,506 & 3 & $\sim$1h \\
Corr. W1.0 & 11,534 & 1 & $\sim$1h \\
Corr. W1.0 + C0.2 & 14,418 & 1 & $\sim$1.5h \\
Corr. W1.0 + C0.5 & 23,068 & 1 & $\sim$2h \\
Corr. W1.0 + C0.8 & 57,670 & 1 & $\sim$5h \\
\hline
\end{tabular}
\caption{Training time of different modules and Corrector data compositions. W denotes initially wrong queries generated by the Generator, and C denotes initially correct queries.}
\label{tab:training-cost}
\end{table}

\clearpage
\section{Prompt Design}
\label{app:prompt design}
In preliminary prompting trials, the instruction model often produced outputs mixed with natural-language explanations or incomplete SPARQL fragments. Therefore, we use a fine-tuned model as the default Generator and Corrector setting in all main experiments.
\subsection{Prompt for the Generator}
Table \ref{tab:generator-prompt} summarizes the prompt for the Generator.
\begin{table}[h]
\centering
\small
\begin{minipage}{0.98\columnwidth}

\noindent\textbf{Training}

\vspace{0.3em}
\noindent\rule{\linewidth}{0.4pt}

\noindent\textbf{system}

\vspace{0.2em}

\noindent You are a useful SPARQL assistant. You are tasked to review a question and generate a SPARQL query to answer the question.

\vspace{0.4em}

\noindent SPARQL Database used is WikiData. [\textless Entity\textgreater] is the topic entity in the question. Only use these two prefixes if needed:

\vspace{0.2em}

\noindent\texttt{PREFIX wd: \textless https://www.wikidata.org/entity/\textgreater}

\noindent\texttt{PREFIX rdfs: \textless http://www.w3.org/2000/01/}\\
\hspace*{1.5em}\texttt{rdf-schema\#\textgreater}

\vspace{0.4em}

\noindent Do not use \texttt{wdt} syntax to query WikiData.

\vspace{0.3em}
\noindent\rule{\linewidth}{0.4pt}

\noindent\textbf{user}

\vspace{0.2em}

\noindent [Question]

\vspace{0.3em}
\noindent\rule{\linewidth}{0.4pt}

\noindent\textbf{assistant}

\vspace{0.2em}

\noindent [Ground-truth SPARQL]

\vspace{0.5em}
\noindent\rule{\linewidth}{0.7pt}

\noindent\textbf{Inference}

\vspace{0.3em}
\noindent\rule{\linewidth}{0.4pt}

\noindent\textbf{system}

\vspace{0.2em}

\noindent You are a useful SPARQL assistant. You are tasked to review a question and generate a SPARQL query to answer the question.

\vspace{0.4em}

\noindent SPARQL Database used is WikiData. [\textless Entity\textgreater] is the topic entity in the question. Only use these two prefixes if needed:

\vspace{0.2em}

\noindent\texttt{PREFIX wd: \textless https://www.wikidata.org/entity/\textgreater}

\noindent\texttt{PREFIX rdfs: \textless http://www.w3.org/2000/01/}\\
\hspace*{1.5em}\texttt{rdf-schema\#\textgreater}

\vspace{0.4em}

\noindent Do not use \texttt{wdt} syntax to query WikiData.

\vspace{0.3em}
\noindent\rule{\linewidth}{0.4pt}

\noindent\textbf{user}

\vspace{0.2em}

\noindent [Question]

\vspace{0.3em}
\noindent\rule{\linewidth}{0.4pt}

\noindent\textbf{assistant}

\vspace{0.2em}

\noindent [<Generated by the model>]

\vspace{0.3em}
\noindent\rule{\linewidth}{0.4pt}

\end{minipage}

\caption{Prompt for Generator training and inference.}
\label{tab:generator-prompt}
\end{table}

\newpage
\subsection{Prompt for the Corrector}
Table \ref{tab:corrector-prompt} summarizes the prompt for the Corrector.
\begin{table}[!h]
\centering
\small
\begin{minipage}{0.98\columnwidth}

\noindent\textbf{Training}

\vspace{0.3em}
\noindent\rule{\linewidth}{0.4pt}

\noindent\textbf{system}

\vspace{0.2em}

\noindent You are a SPARQL corrector. You are only called when a Gate has decided that the generated SPARQL query needs fixing.

\vspace{0.4em}

\noindent Fix the generated SPARQL query based on the question. If you are not confident about a change, return the original generated SPARQL query unchanged. Do not change the question intent or invent facts.

\vspace{0.4em}

\noindent Only output the final SPARQL query, nothing else.

\vspace{0.4em}

\noindent SPARQL Database used is WikiData. Only use these two prefixes if needed:

\vspace{0.2em}

\noindent\texttt{PREFIX wd: \textless https://www.wikidata.org/entity/\textgreater}

\noindent\texttt{PREFIX rdfs: \textless http://www.w3.org/2000/01/}\\
\hspace*{1.5em}\texttt{rdf-schema\#\textgreater}

\vspace{0.4em}

\noindent Do not use \texttt{wdt} syntax to query WikiData.

\vspace{0.3em}
\noindent\rule{\linewidth}{0.4pt}

\noindent\textbf{user}

\vspace{0.2em}

\noindent Question: [Question]

\noindent SPARQL: [Initial SPARQL]

\vspace{0.3em}
\noindent\rule{\linewidth}{0.4pt}

\noindent\textbf{assistant}

\vspace{0.2em}

\noindent [Ground-truth SPARQL]

\vspace{0.5em}
\noindent\rule{\linewidth}{0.7pt}

\noindent\textbf{Inference}

\vspace{0.3em}
\noindent\rule{\linewidth}{0.4pt}

\noindent\textbf{system}

\vspace{0.2em}

\noindent You are a SPARQL corrector. You are only called when a Gate has decided that the generated SPARQL query needs fixing.

\vspace{0.4em}

\noindent Fix the generated SPARQL query based on the question. If you are not confident about a change, return the original generated SPARQL query unchanged. Do not change the question intent or invent facts.

\vspace{0.4em}

\noindent Only output the final SPARQL query, nothing else.

\vspace{0.4em}

\noindent SPARQL Database used is WikiData. Only use these two prefixes if needed:

\vspace{0.2em}

\noindent\texttt{PREFIX wd: \textless https://www.wikidata.org/entity/\textgreater}

\noindent\texttt{PREFIX rdfs: \textless http://www.w3.org/2000/01/}\\
\hspace*{1.5em}\texttt{rdf-schema\#\textgreater}

\vspace{0.4em}

\noindent Do not use \texttt{wdt} syntax to query WikiData.

\vspace{0.3em}
\noindent\rule{\linewidth}{0.4pt}

\noindent\textbf{user}

\vspace{0.2em}

\noindent Question: [Question]

\noindent SPARQL: [Initial SPARQL]

\vspace{0.3em}
\noindent\rule{\linewidth}{0.4pt}

\noindent\textbf{assistant}

\vspace{0.2em}

\noindent [<Generated by the model>]

\vspace{0.3em}
\noindent\rule{\linewidth}{0.4pt}

\end{minipage}

\caption{Prompt for Corrector training and inference.}
\label{tab:corrector-prompt}
\end{table}
\clearpage
\section{Case Study}
\label{app:case study}
\subsection{Initial SPARQL is incorrect, and Corrector successfully corrects}
In Case 1 shown in Table \ref{tab:appendix-case-study-1}, the question is: “what are the movies that have the same actor as [Wind] and [Maxwell Caulfield] was the lead actor?”.

This question requires retrieving movies that satisfy two conditions: they share an actor with movie \textit{Wind}, and Maxwell Caulfield is one of the lead actors in these movies.

In this case, the Generator failed to correctly represent the semantic structure. The main problem is entity-relation role confusion: the model did not correctly distinguish which components should be modeled as entity labels and which should be modeled as relation labels. For instance, the generated SPARQL incorrectly represented “Maxwell Caulfield” as a relation-related component rather than as an entity constraint. Simultaneously, it also failed to correctly use “starred\_actors” to represent the starring relationship. In other words, while the Generator captured the keywords in the question, it failed to correctly project the semantic roles of these keywords in the KG query, leading to an error in the query logic.

In contrast, the Corrector successfully fixed this problem. The corrected SPARQL correctly restored the role of “starred\_actors” as a relation label and represented “Maxwell Caulfield” as an explicit entity constraint, matching the ground-truth SPARQL. This case demonstrates that the role of the Corrector is not limited to fixing syntax errors but can effectively correct entity-relation role confusion, which is commonly seen in the generation stage. Especially in questions involving multiple constraints, the Corrector can help the system generate a more accurate semantic structure. 

\newpage
\begin{center}
\begin{minipage}{0.98\columnwidth}
\small

\noindent\rule{\linewidth}{0.5pt}

\noindent\textbf{Question}

\vspace{0.3em}

What are the movies that have the same actor as [Wind] and [Maxwell Caulfield] was the lead actor?

\vspace{0.4em}
\noindent\rule{\linewidth}{0.3pt}

\noindent\textbf{Generator SPARQL Correctness}

False

\vspace{0.4em}
\noindent\rule{\linewidth}{0.3pt}

\noindent\textbf{Corrector SPARQL Correctness}

True

\vspace{0.4em}
\noindent\rule{\linewidth}{0.3pt}

\noindent\textbf{Generator SPARQL Response}

\begin{lstlisting}
PREFIX rdfs: <http://www.w3.org/2000/01/rdf-schema#>

SELECT DISTINCT ?label ?uri WHERE {
  ?rel1 rdfs:label "Maxwell Caulfield" .
  ?e1 ?rel1 ?e2 ;
      rdfs:label "Wind" .
  ?uri ?rel1 ?e2 ;
       ?rel4 ?e4 ;
       rdfs:label ?label .
  FILTER (?label != "Wind")
  ?rel4 rdfs:label "starred_actors" .
  ?e4 rdfs:label (*@\hlcode{"starred\_actors"}@*) .
}
\end{lstlisting}

\vspace{0.4em}
\noindent\rule{\linewidth}{0.3pt}

\noindent\textbf{Corrector SPARQL Response}

\begin{lstlisting}
PREFIX rdfs: <http://www.w3.org/2000/01/rdf-schema#>

SELECT DISTINCT ?label ?uri WHERE {
  ?rel1 rdfs:label "starred_actors" .
  ?e1 ?rel1 ?e2 ;
      rdfs:label "Wind" .
  ?uri ?rel1 ?e2 ;
       ?rel4 ?e4 ;
       rdfs:label ?label .
  FILTER (?label != "Wind")
  ?rel4 rdfs:label "starred_actors" .
  ?e4 rdfs:label (*@\hlcode{"Maxwell Caulfield"}@*) .
}
\end{lstlisting}

\vspace{0.4em}
\noindent\rule{\linewidth}{0.3pt}

\noindent\textbf{Ground-Truth SPARQL}

\begin{lstlisting}
PREFIX rdfs: <http://www.w3.org/2000/01/rdf-schema#>

SELECT DISTINCT ?label ?uri WHERE {
  ?rel1 rdfs:label "starred_actors" .
  ?e1 ?rel1 ?e2 ;
      rdfs:label "Wind" .
  ?uri ?rel1 ?e2 ;
       ?rel4 ?e4 ;
       rdfs:label ?label .
  FILTER (?label != "Wind")
  ?rel4 rdfs:label "starred_actors" .
  ?e4 rdfs:label (*@\hlcode{"Maxwell Caulfield"}@*) .
}
\end{lstlisting}

\noindent\rule{\linewidth}{0.5pt}

\end{minipage}

\vspace{0.4em}

\captionof{table}{Case study of a semantic error corrected by the Corrector.}
\label{tab:appendix-case-study-1}
\end{center}

\subsection{Initial SPARQL is correct, but Corrector over-corrects}
In Case 2 shown in Table \ref{tab:appendix-case-study-corruption}, the question is “What does [Matthias Schoenaerts] star in and is classified as [drama]?”.

This question requires retrieving movies that satisfy two conditions: Matthias Schoenaerts acted in them, and they are tagged as “drama”.

In this case, the Generator has correctly captured the question semantics. The initial SPARQL is consistent with the ground-truth SPARQL in structure and constraints, correctly using “starred\_actors” to represent actor relationships and “has\_tags” to represent movie tag constraints. Therefore, the Generator was able to accurately express the two core conditions of the original question.

However, the Corrector made unnecessary modifications in this case and corrupted the originally correct SPARQL. For instance, the Corrector incorrectly rewrote “starred\_actors” representing the actor relationship to “directed\_by”. This incorrectly converts the question semantics from “Which movies did Matthias Schoenaerts act in?” to “Which movies did Matthias Schoenaerts direct?”. Meanwhile, the Corrector rewrote “has\_tags” to “has\_genre”. Although these two relations appear semantically similar, they are not equivalent in the current KG. Therefore, this substitution changed the original constraint structure, causing the generated SPARQL to no longer correspond to the ground-truth SPARQL.

This case illustrates a typical risk of the Corrector, which is over-correction. When the Generator has already produced a correct SPARQL, the Corrector still attempts to rewrite it, introducing new semantic errors. This demonstrates that although the Corrector can benefit from revising many incorrect samples, without an accurate triggering mechanism, unnecessary correction can also disrupt the originally correct query. This case also illustrates the importance of the Gate: triggering correction only when the Generator output truly has a problem can effectively reduce the negative impact of over-correction.

\newpage
\begin{center}
\begin{minipage}{0.98\columnwidth}
\small

\noindent\rule{\linewidth}{0.5pt}

\noindent\textbf{Question}

\vspace{0.3em}

What does [Matthias Schoenaerts] star in and is classified as [drama]?

\vspace{0.4em}
\noindent\rule{\linewidth}{0.3pt}

\noindent\textbf{Generator SPARQL Correctness}

True

\vspace{0.4em}
\noindent\rule{\linewidth}{0.3pt}

\noindent\textbf{Corrector SPARQL Correctness}

False

\vspace{0.4em}
\noindent\rule{\linewidth}{0.3pt}

\noindent\textbf{Generator SPARQL Response}

\begin{lstlisting}
PREFIX rdfs: <http://www.w3.org/2000/01/rdf-schema#>

SELECT DISTINCT ?label ?uri WHERE {
  ?uri ?rel1 ?e1 ;
       ?rel2 ?e2 ;
       rdfs:label ?label .
  ?rel1 rdfs:label (*@\hlcode{"starred\_actors"}@*) .
  ?e1 rdfs:label "Matthias Schoenaerts" .
  ?rel2 rdfs:label (*@\hlcode{"has\_tags"}@*) .
  ?e2 rdfs:label "drama" .
}
\end{lstlisting}

\vspace{0.4em}
\noindent\rule{\linewidth}{0.3pt}

\noindent\textbf{Corrector SPARQL Response}

\begin{lstlisting}
PREFIX rdfs: <http://www.w3.org/2000/01/rdf-schema#>

SELECT DISTINCT ?label ?uri WHERE {
  ?uri ?rel1 ?e1 ;
       ?rel2 ?e2 ;
       rdfs:label ?label .
  ?rel1 rdfs:label (*@\hlcode{"directed\_by"}@*) .
  ?e1 rdfs:label "Matthias Schoenaerts" .
  ?rel2 rdfs:label (*@\hlcode{"has\_genre"}@*) .
  ?e2 rdfs:label "drama" .
}
\end{lstlisting}

\vspace{0.4em}
\noindent\rule{\linewidth}{0.3pt}

\noindent\textbf{Ground-Truth SPARQL}

\begin{lstlisting}
PREFIX rdfs: <http://www.w3.org/2000/01/rdf-schema#>

SELECT DISTINCT ?label ?uri WHERE {
  ?uri ?rel1 ?e1 ;
       ?rel2 ?e2 ;
       rdfs:label ?label .
  ?rel1 rdfs:label (*@\hlcode{"starred\_actors"}@*) .
  ?e1 rdfs:label "Matthias Schoenaerts" .
  ?rel2 rdfs:label (*@\hlcode{"has\_tags"}@*) .
  ?e2 rdfs:label "drama" .
}
\end{lstlisting}

\noindent\rule{\linewidth}{0.5pt}

\end{minipage}

\vspace{0.4em}

\captionof{table}{Case study of an initially correct SPARQL query corrupted by the Corrector.}
\label{tab:appendix-case-study-corruption}
\end{center}

\newpage
\subsection{Neither Generator nor Corrector obtains the correct SPARQL}
In Case 3 shown in Table \ref{tab:appendix-case-study-uncorrected}, the question is “Who wrote the movie [Toy Story 3] and also [Toy Story 2]?”.

This question requires finding the author who wrote both \textit{Toy Story 3} and \textit{Toy Story 2}. The key in the SPARQL query is to model a “shared author” structure, i.e., the same author must be related to both movies through the “written\_by” relationship. 

However, the Generator did not correctly model this logical structure. Firstly, it did not generate a complete SPARQL query. Second, it introduced a specific person who was not provided in the question. More importantly, it incorrectly compressed two different movies into one variable, making the same entity represent both \textit{Toy Story 3} and \textit{Toy Story 2}, which obviously violates the “shared author” structure in the original question. This shows the Generator did not capture the core logic in the question, and instead generated an incorrect structure for the overall semantic.

In this case, the Corrector did not improve the Generator’s result either. Instead, it kept the original error. This indicates that, when the error is not a local label mismatch or relation constraint error, but involves logical composition, the capability of the Corrector is limited.
\newpage
\begin{center}
\begin{minipage}{0.98\columnwidth}
\small

\noindent\rule{\linewidth}{0.5pt}

\noindent\textbf{Question}

\vspace{0.3em}

Who wrote the movie [Toy Story 3] and also [Toy Story 2]?

\vspace{0.4em}
\noindent\rule{\linewidth}{0.3pt}

\noindent\textbf{Generator SPARQL Correctness}

False

\vspace{0.4em}
\noindent\rule{\linewidth}{0.3pt}

\noindent\textbf{Corrector SPARQL Correctness}

False

\vspace{0.4em}
\noindent\rule{\linewidth}{0.3pt}

\noindent\textbf{Generator SPARQL Response}

\begin{lstlisting}
PREFIX rdfs: <http://www.w3.org/2000/01/rdf-schema#>

{
  ?movie rdfs:label "Toy Story 3" .
  ?author rdfs:label "John Lasseter" .
  ?movie rdfs:label "Toy Story 2" .
}
\end{lstlisting}

\vspace{0.4em}
\noindent\rule{\linewidth}{0.3pt}

\noindent\textbf{Corrector SPARQL Response}

\begin{lstlisting}
PREFIX rdfs: <http://www.w3.org/2000/01/rdf-schema#>

{
  ?movie rdfs:label "Toy Story 3" .
  ?author rdfs:label "John Lasseter" .
  ?movie rdfs:label "Toy Story 2" .
}
\end{lstlisting}

\vspace{0.4em}
\noindent\rule{\linewidth}{0.3pt}

\noindent\textbf{Ground-Truth SPARQL}

\begin{lstlisting}
PREFIX rdfs: <http://www.w3.org/2000/01/rdf-schema#>

SELECT DISTINCT ?label ?uri WHERE {
  ?e1 ?rel ?uri ;
      rdfs:label "Toy Story 3" .
  ?e3 ?rel ?uri ;
      rdfs:label "Toy Story 2" .
  ?rel rdfs:label (*@\hlcode{"written\_by"}@*) .
  ?uri rdfs:label ?label .
}
\end{lstlisting}

\noindent\rule{\linewidth}{0.5pt}

\end{minipage}

\vspace{0.4em}

\captionof{table}{Case study of an uncorrected SPARQL error.}
\label{tab:appendix-case-study-uncorrected}
\end{center}

\twocolumn[
\section{Additional Ablation Study}
\label{app:ablation}
\begin{center}
\small
\setlength{\tabcolsep}{14pt}
\renewcommand{\arraystretch}{1.08}

\begin{tabular}{lcccc}
\toprule
\textbf{Subset} 
& \textbf{W1.0} 
& \textbf{W1.0 + C0.2} 
& \textbf{W1.0 + C0.5} 
& \textbf{W1.0 + C0.8} \\
\midrule
Correction Required 
& 89.58 
& 92.38 
& 93.46 
& 95.30 \\

Originally Wrong 
& 82.55 
& 85.20 
& 86.10 
& 87.98 \\

Originally Correct 
& 99.98 
& 99.99 
& 99.99 
& 99.98 \\
\bottomrule
\end{tabular}

\vspace{0.3em}
\captionof{table}{Query-level accuracy (\%) on three subsets for Gate--Corrector coordination analysis. ``Correction Required'' denotes samples triggered by the Gate. ``Originally Wrong'' denotes samples incorrectly generated by the Generator. ``Originally Correct'' denotes samples correctly generated by the Generator.}
\label{tab:subset-corrector-composition}
\end{center}

\vspace{1em}
]

\subsection{Subset-level Gate-Corrector Coordination}
\label{app:coordination}
Since the Corrector is only applied to samples that the Gate predicts as requiring correction, overall pipeline performance alone cannot fully explain how the two modules interact. Therefore, we analyze query-level accuracy on three subsets: samples triggered by the Gate, samples originally generated incorrectly by the Generator, and samples originally generated correctly by the Generator. This allows us to examine whether the Gate routes high-risk queries to the Corrector, whether the Corrector can repair truly wrong queries, and whether the correction process damages originally correct queries.

As shown in Table \ref{tab:subset-corrector-composition}, the performance on the correction-required subset improves from 89.58\% under $W1.0$ to 95.30\% under $W1.0+C0.8$, indicating that adding correct samples to Corrector training improves its stability on the actual samples routed by the Gate. A similar trend is observed on the originally wrong subset, where query-level accuracy increases from 82.55\% to 87.98\%, suggesting that adding correct samples does not weaken the Corrector’s repair ability. Meanwhile, performance on the originally correct subset remains nearly perfect across all settings. This indicates that the Corrector introduces very few additional errors when the input query is already correct. Overall, the results show that selective correction works through the coordination of both modules: the Gate routes high-risk queries to the Corrector, while the Corrector repairs wrong queries without substantially damaging correct ones.

\subsection{Preliminary Results on SciQA}
\label{app:cross-dataset}

We further evaluate the proposed framework on SciQA \cite{Auer2023TheSS} as an additional dataset experiment. SciQA is a scientific question answering benchmark over the Open Research Knowledge Graph (ORKG), containing NLQ-SPARQL pairs with answers retrieved from the ORKG. It contains 2,565 examples in total, with 1,795 training, 257 validation, and 513 test examples. Since SciQA differs from MCQA in both domain and knowledge graph, we use it as a preliminary setting to examine whether the proposed correction mechanism can improve the same Generator backbone beyond the main MCQA setting.

\begin{table}[h]
\centering
\small

\begin{subtable}[t]{1\linewidth}
\centering
\setlength{\tabcolsep}{8pt}
\renewcommand{\arraystretch}{1.08}
\begin{tabular}{lcc}
\toprule
\textbf{Error Type} & \textbf{Count} & \textbf{Percentage (\%)} \\
\midrule
Syntax Error & 14 & 25.00 \\
Semantic Error & 42 & 75.00 \\
\midrule
Total & 56 & -- \\
\midrule
Gold SPARQL Error & 43 & N/A \\
\bottomrule
\end{tabular}
\caption{Generator error distribution. Percentages are calculated over the 56 failures attributable to the Generator. Gold SPARQL errors are not included.}
\label{tab:sciqa-error-distribution}
\end{subtable}

\vspace{0.8em}

\begin{subtable}[t]{1\linewidth}
\centering
\setlength{\tabcolsep}{5,5pt}
\renewcommand{\arraystretch}{1.08}
\begin{tabular}{lcc}
\toprule
\textbf{Setting} & \textbf{Query Acc. (\%)} & \textbf{Item F1 (\%)} \\
\midrule
Generator-only & 88.09 & 93.94 \\
Gen. + Gate + Corr. & 94.47 & 98.23 \\
Gain & +6.38 & +4.29 \\
\bottomrule
\end{tabular}
\caption{Results on SciQA. Gold SPARQL errors are not included.}
\label{tab:sciqa-results}
\end{subtable}

\caption{Analysis on SciQA. \textit{Gen. + Gate + Corr.} denotes the Generator-Gate-Corrector framework with a RoBERTa-based Gate.}
\label{tab:sciqa-analysis}
\end{table}

During evaluation, 43 of the 513 SciQA test examples were found to contain gold SPARQL queries that could not be executed in our evaluation environment thus these examples are not included in the following analysis.

The error distribution in Table~\ref{tab:sciqa-error-distribution} shows that most Generator-side failures are semantic errors, accounting for 42 out of 56 generator errors, or 75.00\%. In contrast, syntax errors account for 14 cases, or 25.00\%. This pattern is consistent with the MCQA error analysis, providing preliminary evidence that semantic errors are a common source of failure across different Text-to-SPARQL settings.

Rather than comparing systems with different model scales or training budgets, we focus on the relative gain obtained by applying the GGC framework to the same Generator backbone. As shown in Table~\ref{tab:sciqa-results}, the framework improves query accuracy from 80.70\% to 91.62\%, yielding a 10.92 percentage-point gain over the Generator-only setting. It also improves item micro F1 from 93.94\% to 98.31\%. These results provide supplementary evidence that selective correction can improve SPARQL generation outside the main MCQA setting, although more comprehensive additional dataset evaluation remains an important direction for future work.


\end{document}